\documentclass[conference]{IEEEtran}
\IEEEoverridecommandlockouts
\usepackage{cite}
\usepackage{amsmath,amssymb,amsfonts}
\usepackage{algorithmic}
\usepackage{graphicx}
\usepackage{textcomp}
\usepackage{booktabs}
\usepackage{multirow}
\usepackage{xcolor}
\usepackage{url}
\usepackage{subcaption}
\usepackage{enumitem} 
\usepackage{acronym}
\newacro{cap}[CaP]{Code-as-Policy}
\newacro{aasp}[AasP]{Agent-as-Policy}
\newacro{rap}[RAP]{Robotic Agent as Policy}
\newacro{alrm}[ALRM]{Agentic LLM for Robot Manipulation}
\newacro{tap}[TaP]{Tool-as-Policy}
\newacro{llm}[LLM]{Large Language Model}
\newacro{vlm}[VLM]{Vision Language Model}
\newacro{vla}[VLA]{Vision Language Action}
\newacroplural{llm}[LLMs]{Large Language Models}
\newacro{ros}[ROS]{Robotic Operating System}
\newacro{api}[API]{Application Programmable Interface}
\newacroplural{api}[APIs]{Application Programmable Interfaces}

\def\BibTeX{{\rm B\kern-.05em{\sc i\kern-.025em b}\kern-.08em
    T\kern-.1667em\lower.7ex\hbox{E}\kern-.125emX}}
\begin{document}

\title{ALRM: Agentic LLM for Robotic Manipulation\\
}

\author{
\IEEEauthorblockN{ Vitor Gaboardi dos Santos\textsuperscript{1}\textsuperscript{2}, 
 Ibrahim Khadraoui\textsuperscript{1}, 
 Ibrahim Farhat\textsuperscript{1},  
 Hamza Yous\textsuperscript{1}, 
 Samy Teffahi\textsuperscript{1},
 Hakim Hacid\textsuperscript{1}}
\IEEEauthorblockA{\textsuperscript{1}Technology Innovation Institute, Abu Dhabi, UAE  \\
\textsuperscript{2}Dublin City University, Dublin, Ireland\\
Brahim.Farhat@tii.ae\textsuperscript{1} and vitor.gaboardidossantos2@mail.dcu.ie\textsuperscript{2}}
}

\maketitle

\begin{abstract}
\acp{llm} have recently empowered agentic frameworks to exhibit advanced reasoning and planning capabilities. However, their integration in robotic control pipelines remains limited in two aspects: (1) prior \ac{llm}-based approaches often lack modular, agentic execution mechanisms, limiting their ability to plan, reflect on outcomes, and revise actions in a closed-loop manner; and (2) existing benchmarks for manipulation tasks focus on low-level control and do not systematically evaluate multistep reasoning and linguistic variation. In this paper, we propose \ac{alrm}, an \ac{llm}-driven agentic framework for robotic manipulation. \ac{alrm} integrates policy generation with agentic execution through a ReAct-style reasoning loop, supporting two complementary modes: \ac{cap} for direct executable control code generation, and \ac{tap} for iterative planning and tool-based action execution. To enable systematic evaluation, we also introduce a novel simulation benchmark comprising 56 tasks across multiple environments, capturing linguistically diverse instructions. Experiments with ten \acp{llm} demonstrate that \ac{alrm} provides a scalable, interpretable, and modular approach for bridging natural language reasoning with reliable robotic execution. Results reveal Claude-4.1-Opus as the top closed-source model and Falcon-H1-7B as the top open-source model under \ac{cap}. Data and code is available on \url{https://tiiuae.github.io/ALRM} \\
\vspace{0.5em}

\end{abstract}


\acresetall

\section{Introduction}

Robotic systems have traditionally been constructed around explicit, pre-programmed task specifications, with control logic encoded as manually scripted behaviors \cite{ajaykumar2021survey}. These systems rely on a multi-layered software stack, ranging from operating systems like \ac{ros}, to hardware abstraction \acp{api}, and middleware such as MoveIt, to orchestrate perception, planning, and actuation \cite{macenski2022robot}. While decades of research have yielded highly optimized modules for control and perception, these components remain rigid, domain-specific, and often incapable of generalizing to novel tasks or adapting dynamically to environmental changes \cite{zhang2025generative}.

Recent advances in \acp{llm} have unlocked powerful capabilities across a broad range of domains \cite{zhao2023survey}. Beyond natural language generation, modern \acp{llm} can perform code synthesis, image interpretation, tool use, and symbolic reasoning—all from natural language prompts \cite{zhao2023survey}. These multi-modal, general-purpose capabilities position \acp{llm} as a compelling foundation for embodied AI, where perception, reasoning, and action must be integrated within dynamic, real-world environments \cite{Wang_etal_LLM3_2024}.

In the context of robotics, \acp{llm} bring the potential to interpret unstructured language, reason over symbolic goals, and interface with APIs to automate control. Emerging frameworks such as Code-as-Policy \cite{Liang_etal_CodeAsPolicies_2022} position \acp{llm} as policy generators that translate high-level goals into executable code, enabling zero-shot task automation without retraining. Additionally, \acp{llm} assist with low-level control grounding and abstract task decomposition \cite{Ahn_etal_SayCan_2022, Li_etal_ProgPrompt_2022}, bridging human intent and robot behavior. However, early efforts integrating \acp{llm} into robotics are often limited by static planning, lack of interpretability, and an absence of closed-loop mechanisms for reflection during task execution \cite{kim2024survey}.

In parallel, the field of agentic AI has demonstrated the effectiveness of architectures that plan, act, and refine. Frameworks such as ReAct \cite{react} and Reflexion \cite{shinn2023reflexion} demonstrate how structured reasoning can be combined with interactive execution, enabling systems to decompose tasks, invoke tools or generate code, and adapt actions based on feedback. 


Furthermore, although simulation benchmarks for robotic manipulation exist \cite{liu2023libero, james2020rlbench, robocasa2024}, they often lack tasks that require reasoning over the environment or objects, and tasks that involve multiple sequential steps. In addition, most benchmarks do not systematically evaluate linguistic variation, leaving task instructions untested under different wordings, syntactic structures, or reasoning levels. These limitations constrain the assessment of LLM-based agents in reasoning and executing more complex, multistep tasks. 


In this paper, we introduce \textbf{\ac{alrm}}, a novel framework that combines LLM-based action generation with agentic execution for robotic manipulation. \ac{alrm} is built around an LLM-based agent architecture, where specialized agents handle task planning and execution. It also uses ReAct-style loop \cite{react}, enabling the system to iteratively reason about goals, monitor execution, and revise plans as needed. The framework supports two complementary modes of operation: \ac{cap} and \ac{tap}. In \ac{cap} mode, actions are executed by directly running executable control code, while in \ac{tap} mode, actions are executed via tool calling capabilities of \acp{llm}.

To evaluate \ac{alrm}, we present a benchmarking framework designed to address limitations of prior robotic manipulation benchmarks. Our benchmark emphasizes high-level, multistep tasks with different levels of complexity, from simple lexical variations to complex reasoning, providing a structured testbed for assessing the execution capabilities of \ac{llm}-based agents in robotic manipulation.

In summary, we make the following key contributions:
\begin{itemize}
    \item We propose \textbf{\ac{alrm}}, an \ac{llm}-driven agentic framework for robotic planning and execution that supports two modes of operation: \ac{cap} and \ac{tap}.    

    \item We introduce a novel benchmark of 56 instructions for robotic manipulation that emphasizes multistep, high-level tasks with rich linguistic diversity. 

    \item We conduct a systematic evaluation of ten \acp{llm}, reporting task success rate and execution latency across both \ac{cap} and \ac{tap} execution modes.

    \item We develop a reusable simulation environment based on \textbf{Gazebo}, integrated with \ac{ros} and MoveIt to support deployment and testing of \ac{llm} driven manipulation.
    
\end{itemize}

Experiments show that Claude-4.1-Opus was the best closed-source model across both TaP and CaP modes, achieving average success rates of 93.5\% and 92.6\%, respectively. On the other hand, Falcon-H1-7B was the best-performing open-source model under CaP, achieving the same 84.3\% success rate as DeepSeek-V3.1 while requiring less than half the latency. These results demonstrate the ability of \acp{llm} to solve high-level and abstract robotic manipulation tasks using our proposed agentic framework.

The remainder of this paper is organized as follows: Section~\ref{sec:related_work} reviews related work on leveraging \acp{llm} for robotic control and existing benchmarks for manipulation tasks. Section~\ref{sec:rap} introduces \ac{alrm}, our proposed agentic \ac{llm} framework for robotic manipulation. Section~\ref{sec:benchmark} details our benchmark for evaluating pick-and-place tasks. Section~\ref{sec:experiments} outlines the experimental setup. Section~\ref{sec:results} reports the results of evaluating ten LLMs on our benchmark. Finally, Section~\ref{sec:conclusion} concludes the paper and discusses future research directions.

\section{Related Work}
\label{sec:related_work}

\begin{figure*}[t]
\centering
\includegraphics[width=1\textwidth]{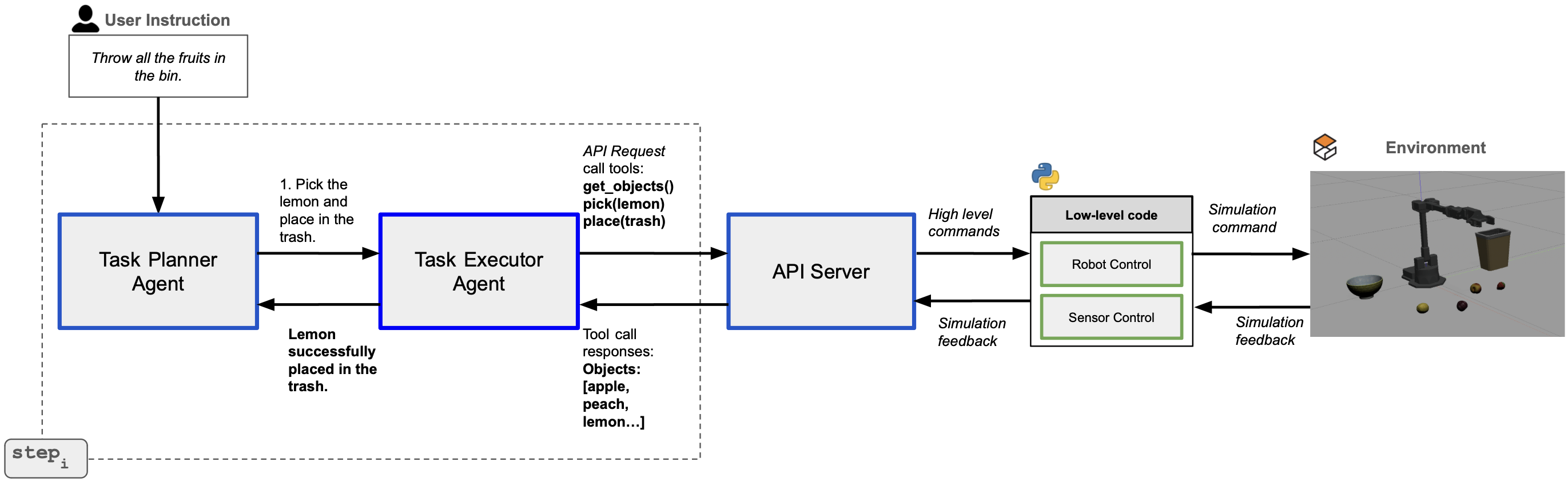}
\caption{Proposed LLM-based agent architecture for solving high-level robotic arm manipulation tasks. The architecture consists of three main modules: (1) Task Planner Agent, (2) Task Executor Agent, and (3) API Server.}
\label{fig_approach_overview}
\end{figure*}

In this section, we review prior work on using \acp{llm} for robotic control, which can be broadly categorized into \emph{explicit} and \emph{implicit} control approaches. We also discuss existing benchmarks for evaluating robotic manipulation tasks.

\emph{\textbf{Explicit robot control}}.
In the explicit case, the model is trained end-to-end to predict low-level control targets such as end-effector positions or joint states, which are then executed by the robot using inverse kinematics or trajectory tracking. This paradigm, adopted by many \ac{vla} models, directly maps perception and language inputs to continuous motor commands, achieving strong performance but often at the cost of generalization. 

Recent works illustrate the strengths and limitations of this paradigm. \textit{$\pi_0$}~\cite{PhysicalInt_Pi0_2024} introduces a high-frequency \ac{vla} model trained on over 10,000 hours of robotic data. It uses a flow-matching transformer architecture to generate 50 Hz continuous actions directly from language and image inputs, achieving impressive zero-shot generalization. Unlike \textit{$\pi_0$}, \textit{OpenVLA}~\cite{Kim_etal_OpenVLA_2024} proposes an open source approach following a different architecture by combining vision encoders with \ac{llm}-based action decoders. The actions are predicted auto-progressively without the need of adding a separate dedicated parameters as experts. While effective for imitation learning and fine-tuning, it remains bound to closed-loop policies without explicit reasoning. More recently, \textit{Gemini Robotics}~\cite{Google_GeminiRobotics_2025} extends explicit policies through large-scale video-pretrained models that transfer across embodiments, highlighting the promise of latent affordances for manipulation. Yet, like prior efforts, it is still constrained by reliance on massive data and lacks mechanisms for reasoning, planning, or agentic reflection. 

\emph{\textbf{Implicit robot control}}.
Implicit control leverages pretrained \acp{llm} to generate code or action sequences that interface with the robot’s control stack. Instead of regressing raw motor commands, the \ac{llm} produces high-level instructions or API calls (e.g. pick, place, move\_to), which are executed by motion planners and inverse kinematics solvers. This direction bypasses the need for training specialized end-to-end policies and enables robots to exploit existing control infrastructure by providing the context and APIs documentations to the models. 

A prominent early effort is \textit{Code-as-Policy (CaP)}~\cite{Liang_etal_CodeAsPolicies_2022}, which demonstrated that \acp{llm} can synthesize perception-conditioned robot policy code from natural language instructions. By leveraging hierarchical code prompting, CaP enables structured API calls and waypoint-based behaviors that generalize to novel tasks, though its one-shot generation strategy limits adaptivity and reflection. To address brittleness in free-form outputs, \textit{ProgPrompt}~\cite{Li_etal_ProgPrompt_2022} introduced structured prompts describing available actions, objects, and example programs. This design grounds LLM outputs in the robot’s capabilities, yielding executable task plans, but still assuming static pipelines without runtime monitoring. Building on these foundations, \textit{LLM$^3$}~\cite{Wang_etal_LLM3_2024} proposed an interface between task and motion planning, incorporating motion-planning feedback so that failed executions can prompt iterative refinement. While this introduces a closed-loop element, the framework remains tailored to task-and-motion scenarios without modular agentic orchestration. \textit{SayCan}~\cite{Ahn_etal_SayCan_2022} emphasized feasibility by coupling \ac{llm}-based instruction parsing with pretrained low-level skills and value functions, filtering actions to ensure contextual appropriateness. SayCan achieved long-horizon task completion in real-world environments but is constrained by its reliance on fixed skill libraries and the absence of explicit symbolic reasoning. 
\textit{VoxPoser}~\cite{huang2023voxposer} combines \ac{llm} inference of constraints and affordances with a \ac{vlm} to generate 3D value maps, which are then converted into dense 6-DoF motion trajectories by external planners. 

In contrast to these efforts, \textbf{\ac{alrm}} introduces a unified and modular agentic framework that integrates both \emph{code generation} and \emph{tool based} execution for robotic control. Our system supports two complementary modes: a \textit{\ac{cap}} approach that directly generates executable control code, and a \textit{\ac{tap}} mode that leverages the tool calling capabilities of LLMs for robot control. Unlike prior work, \ac{alrm} incorporates agent coordination, reflection over execution outcomes, and closed-loop task revision. 

\emph{\textbf{Benchmarks for robotic manipulation systems}}.
Several benchmarks have been proposed for evaluating robotic manipulation systems, but they differ from our work in scope, complexity, and language diversity.
ACRV \cite{leitner2017acrv} is a physical benchmark and evaluation protocol for robotic shelf picking applications. It includes a set of globally available objects and a standardized shelf setup to facilitate reproducibility. However, task definitions are limited, language diversity is scarce, and no simulation environment is provided.
RLBench \cite{james2020rlbench} is a benchmark of over 100 manipulation tasks, ranging from  pick-and-place to stacking wine bottles tasks. Each task includes multiple natural language instructions, although linguistic variation is limited since these mainly involve word replacement or minor rephrasing. 
RoboCasa \cite{robocasa2024} contains 100 tasks for a kitchen environment, including 25 atomic tasks (e.g., single pick-and-place, pressing buttons) and 75 composite tasks generated with LLM guidance (e.g., clear the table, wash dishes). While composite tasks introduce more complexity, task definitions remain straightforward and not linguistic diverse. Additionally, tasks commonly include only objects relevant to the task, which simplifies reasoning about which objects to interact with.


In this paper, we release a novel benchmark that emphasizes high-level manipulation tasks requiring more complex reasoning and at least two sequential pick-and-place actions. Tasks are designed with four levels of difficulty, ranging from simple lexical variations (e.g., replacing words with synonyms) to high-level reasoning requirements. For example, instead of instructing the robot to pick a strawberry and a peach, a task may require the agent to select the two fruits with the lowest calories in the scene. Furthermore, environments contain more objects than required for the task, increasing complexity by forcing the agent to detect and move only relevant objects. Thus, our benchmark provides a foundation for systematically assessing LLMs as agents in robotic manipulation tasks.

\section{LLM-Based Robotic Agent Architecture for Task Planning and Execution}
\label{sec:rap}

Figure \ref{fig_approach_overview} shows the proposed LLM-based agent architecture for solving high-level arm manipulation robotic tasks. The proposed architecture contain three main modules: (1) task planner agent, (2) task executor agent, and (3) API server.

\begin{figure*}[h]
    \centering
    \begin{subfigure}[t]{0.32\textwidth}
        \centering
        \includegraphics[width=\textwidth]{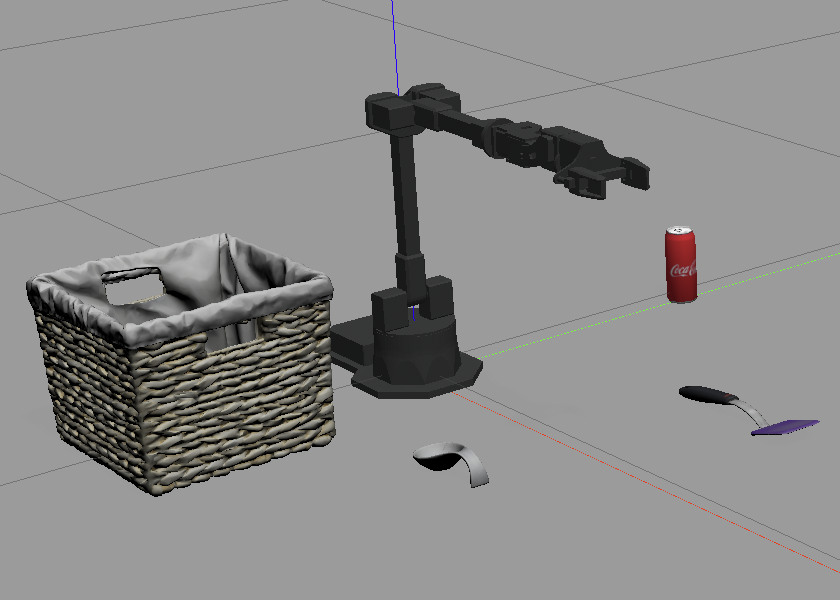}
        \caption{Kitchen utensils environment.}
        \label{fig:sub1}
    \end{subfigure}
    \hfill
    \begin{subfigure}[t]{0.32\textwidth}
        \centering
        \includegraphics[width=\textwidth]{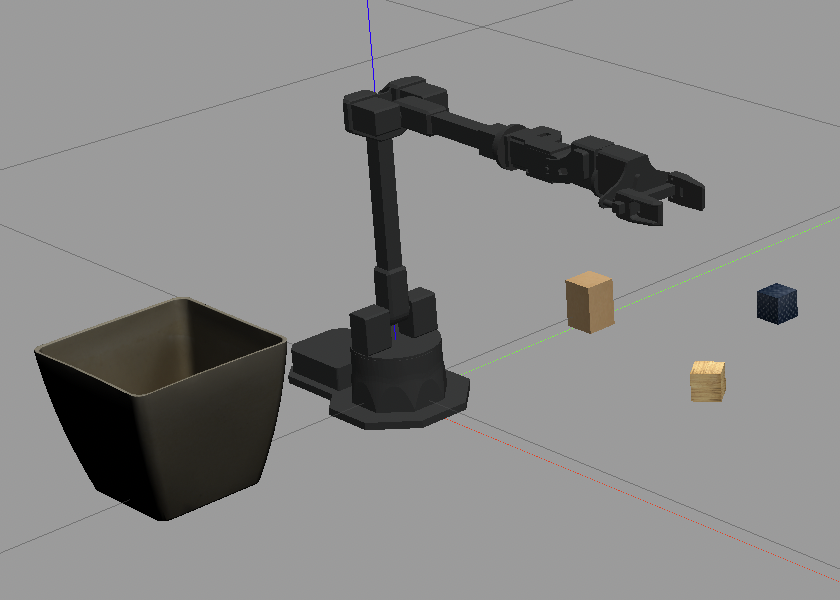}
        \caption{Boxes environment.}
        \label{fig:sub2}
    \end{subfigure}
    \hfill
    \begin{subfigure}[t]{0.32\textwidth}
        \centering
        \includegraphics[width=\textwidth]{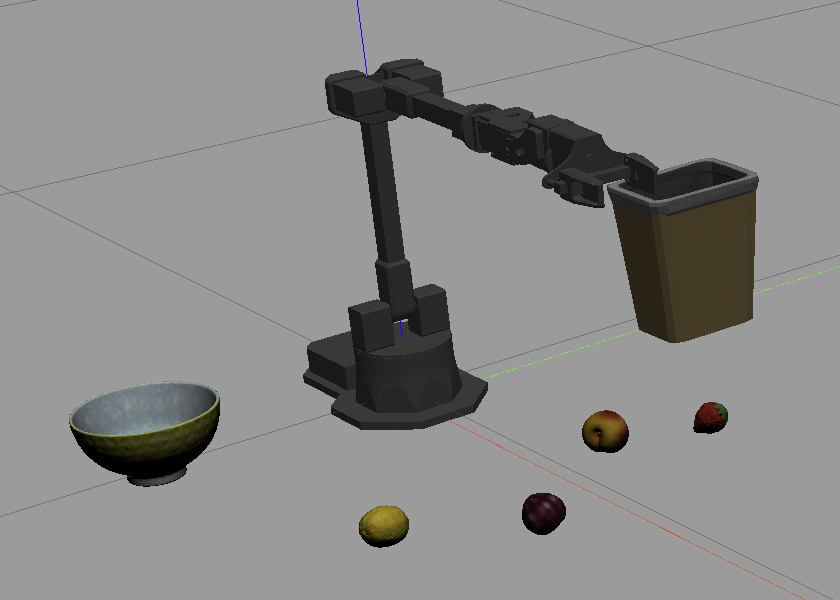}
        \caption{Fruits environment.}
        \label{fig:sub3}
    \end{subfigure}
    \caption{Illustration of the three environments designed for evaluation.}
    \label{fig:environment}
\end{figure*}

The task planner agent incrementally generates a plan through iterative cycles of thought, action, and observation, dynamically creating subtasks based on interactions with the environment and the robot. The task executor agent converts these high-level, natural language subtasks into actions and sends them to the API server, which executes these actions in simulation or on a real robot. After execution, the task executor agent summarizes the results and returns them to the planner in the form of observations, giving context and enabling the planner to reason over the results to generate subsequent actions. These interactive steps continue until the original user task is fulfilled, or it reaches the maximum number of steps. 


\subsection{Task Planner Agent}

The task planner agent is designed based on the ReAct framework~\cite{react}, generating iterative cycles of thought and action to decompose high-level user requests into subtasks. Unlike conventional approaches that rely on pre-defined scripts, the planner leverages LLMs to dynamically generate subtasks, receiving observations from the task executor agent (discussed in the next section) as feedback. This interactive feedback loop enables the planner to adapt to the environment in real time. 

We instruct the LLM to generate subtasks according to high-level templates that cover most manipulation scenarios. Example templates include:

\begin{itemize}
    \item \textit{Get the position of objects in the environment}.
    \item \textit{Get the names of objects present in the environment}.
    \item Execute atomic manipulation tasks such as “\textit{Pick up [\textit{object\_name}] and place it [\textit{spatial relation}] [\textit{destination}].}”
\end{itemize}

By adhering to these templates, the planner can systematically obtain information from the environment and define atomic subtasks for the robotic arm to perform. Additional guidelines ensure that the LLM generates only one action per step and focuses on one object at a time, aligning with the capabilities of a single-arm robot. Moreover, the planner is instructed to generate actions only for objects present in the environment and relevant to the user’s request, preventing the execution of unintended operations.


\subsection{Task Executor Agent}

The task executor agent receives a high-level natural language subtask (e.g., “\textit{Pick the lemon and place it in the trash}”) and generates sequence of actions required to achieve the subtask. The agent has access to a total of eight predefined actions for (1) controlling the robot (\texttt{pick}, \texttt{place}, \texttt{move\_to\_home\_pos}, and \texttt{move\_to}); (2) perceiving the environment (\texttt{get\_objects} and \texttt{get\_reference\_names}); and (3) retrieving poses for grasping or placing (\texttt{compute\_grasp} and \texttt{get\_pose}). These actions can be executed through two complementary modes: \ac{tap} and \ac{cap}.


\subsubsection{Tool-as-Policy (TaP)}

The tool-based executor leverages the tool calling capabilities of LLMs to execute actions as nested tool calls. It operates in interactive steps, where the LLM typically produces a single tool call per step. The result is appended to the conversation history and fed back to the LLM  until the subtask is completed. This design enables the tool-based executor agent to reason over intermediate results and correct mistakes (e.g., referencing an incorrect object name). It also provides greater flexibility, as small errors can be solved without requiring replanning from the task planner agent. However, it increases the number of LLM calls and relies on the model’s ability to reliably perform tool calling.




We provided the \acp{llm} with instructions of best practices for generating tool calls according to high-level subtask templates. Observations returned to the task planner agent are built from the tool call responses, which include descriptive feedback such as “\textit{The lemon was successfully picked}” or “\textit{Object orange not found}”. 


\subsubsection{Code-as-Policy (CaP)}

The code-based executor generates Python code that calls the available functions to fulfill the subtask in a single run. It executes the generated code while maintaining a log of all action executions. This approach is faster than the tool-based executor because the entire subtask is handled at once, requiring fewer \ac{llm} calls and enabling the use of those that do not support function/tool calling. However, it can be more brittle: if any part of the generated code contains an error, the entire subtask execution may fail. Consequently, this approach relies more heavily on the task planner to generate accurate natural language action descriptions.



We provided the executor with instructions of best practices for generating code based on subtask templates, along with definitions of all actions (represented as Python functions), including method descriptions, input parameters, and expected outputs. Furthermore, we provide a one-shot code example for a general pick-and-place task, giving the \ac{llm} a standard reference to generate the code.
Observations returned to the task planner agent are built from the execution logs, and include descriptive feedback similar to the tool-based executor. 



\subsection{API Server for Robotic Control}
We implemented a RESTful API server to control the Interbotix \texttt{wx250s} arm \cite{wx250s_specs} in simulation. The server provides endpoints for pick-and-place, motion commands, and perception queries. Two internal modules connect the API to the robotics backend: \texttt{wx250sRobot}, an abstraction layer that communicates with MoveIt and ROS to control the robot through high-level functions (e.g. \texttt{/pick}, \texttt{/place}), and \texttt{SimPerception}, which retrieves object positions and grasp pose estimates from Gazebo (e.g., \texttt{/compute\_grasp}, \texttt{/get\_reference\_names}). 

This design allows the LLM to focus on generating high-level actions rather than low-level ROS commands, improving performance by reducing the complexity of action generation by the LLM while ensuring generalization to different tasks. At the same time, it provides a clean and reproducible interface for robotic manipulation experiments.


\section{Benchmark}
\label{sec:benchmark}

\begin{table*}[h!]
\centering
\caption{Example of instructions for task 1 of the fruits environment.}
\label{tab:examples_tasks}
\begin{tabular}{cc}
\toprule
\textbf{Paraphrase Category} & \textbf{Instruction} \\
\midrule
Canonical & Pick up the lemon and peach and place them in the trash. \\
Lexical & Grab the lemon and peach and throw them in the garbage. \\
Syntactical & Can you pick up the lemon and peach and place them in the trash? \\
Semantics & Hey! these fruits are rot! toss the lemon and peach in the bin \\
High-level reasoning & Pick up the sourest and the biggest fruits and place them in the bin \\
\bottomrule
\end{tabular}
\end{table*}

To evaluate our proposed method and assess how different \acp{llm} perform in high-level robot manipulation tasks, we designed a benchmark dataset that combines controlled environments, structured tasks, and linguistically diverse instructions.

The benchmark is built around three simulated environments, each containing a robotic manipulator and a set of objects. In each environment, we define three canonical tasks involving multi-object pick-and-place actions. To test the robustness of LLMs to language variation, each canonical task is paraphrased five times, covering six linguistic categories. In total, the benchmark includes 3 environments × 3 tasks × 6 instructions = 54 tasks. We provide ground-truth code and sequences of tool calls for each task, validated in a Gazebo simulation environment. This benchmark is made publicly available to support further evaluation of robotic manipulation planning and execution.



\subsection{Environments}

Figure \ref{fig:environment} shows the three environments developed with different objects for manipulation. In summary, we have the following objects for each environment: 

\begin{enumerate}
    \item Kitchen utensils: spoon, spatula, can of coke, and basket.
    \item Boxes: cardboard box, wooden box, metal box, and container.
    \item Fruits: strawberry, plum, lemon, peach, bowl, and trash.
\end{enumerate}

Each environment challenges \ac{alrm} to reason about object properties and positions in distinct semantic domains.

\subsection{Tasks}

For every environment, we define three canonical tasks, which expresses the task as a direct imperative command (e.g., “\textit{Move the spoon, the coke, and the spatula to the basket}”). Each task involves manipulating at least two objects and may require respecting execution order (e.g., move first a metal box followed by a wooden box), handling multiple reference positions (e.g., in front of the robot, left of the coke can), or taking actions based on object properties (e.g., weight, size, or category). These tasks collectively form a rich dataset in terms of complexity, enabling a comprehensive evaluation of current \ac{llm} agent capabilities.

To evaluate LLM performance beyond the canonical task, we generate five paraphrases per task, each representing a distinct linguistic challenge.

\subsection{Paraphrase Categories}
\label{ref:paraphrase_category}

We adopt four categories, inspired by the paraphrasing literature \cite{dos2024diverse,guo2024benchmarking}, to capture diverse levels of linguistic and reasoning difficulty. Table \ref{tab:examples_tasks} provides examples of paraphrases corresponding to a given canonical instruction. Below, we define each category:

\begin{itemize}
    \item \textbf{Lexical (LEX)}: Replacement of nouns and verbs for synonyms or similar terms.
    \item \textbf{Syntactical (SYN)}: Rephrasing sentences with different syntactic structures (e.g., changing a direct command into a declarative, interrogative, or compound sentence).
    \item \textbf{Semantics (SEM)}: Emphasizes contextual or stylistic changes, including typos, informal phrasing, inclusion of chitchat, or user-specific preferences.
    \item \textbf{High-level reasoning (HLR)}: Abstract instructions that require inference or reasoning to identify objects or define actions.
\end{itemize}

Paraphrases are solved using the same actions of the canonical task without ambiguity. For example, in Environment 3, the sourest fruit must be the lemon regardless of the remaining fruits. Furthermore, although instructions may combine elements from multiple categories, each paraphrase is labeled with a single category representing the core change relative to the canonical task. For instance, a semantics-aware paraphrase may include lexical variations, but it is labeled as semantics.


We generate one paraphrase per category for each task, except for high-level reasoning, which has two paraphrases per task. We prioritize high-level reasoning because it represents more challenging scenarios.

\section{Experimental Setup}
\label{sec:experiments}

In this section, we describe the experimental setup, including the selection of LLMs for evaluation, the evaluation metrics, and the overall evaluation design.

\subsection{Large Language Models}
\label{sub:llms}

We selected the following ten LLMs to evaluate both CaP and TaP modes: 
GPT-5\footnote{\url{https://openai.com/index/introducing-gpt-5/}}, 
Gemini-2.5-Pro\footnote{\url{https://deepmind.google/models/gemini/pro/}}, 
Claude-4.1-Opus\footnote{\url{https://www.anthropic.com/news/claude-opus-4-1}}, 
DeepSeek-V3.1 (non-thinking mode)\footnote{\url{https://api-docs.deepseek.com/news/news250821}}, 
Falcon-H1-7B\footnote{\url{https://huggingface.co/tiiuae/Falcon-H1-7B-Instruct}}, 
Qwen3-8B\footnote{\url{https://huggingface.co/Qwen/Qwen3-8B}}, 
Llama-3.1-8B\footnote{\url{https://huggingface.co/meta-llama/Llama-3.1-8B-Instruct}}, 
DeepSeek-R1-7B\footnote{\url{https://huggingface.co/deepseek-ai/DeepSeek-R1-Distill-Qwen-7B}}, Granite-3.3-8B\footnote{\url{https://huggingface.co/ibm-granite/granite-3.3-8b-base}}, and
Mistral-7B\footnote{\url{https://huggingface.co/mistralai/Mistral-7B-Instruct-v0.2}}. 

We include large-scale LLMs with strong performance on the Berkeley Function Calling Leaderboard \cite{patil2023gorilla} at the time of this work, enabling the evaluation of state-of-the-art solutions for both code and tool-calling generation. We also include small-scale LLMs (up to 8B parameters) to assess whether lighter models can solve robotics applications, since they can be run locally on a single GPU. This selection allows developers to analyze the trade-off between larger models, which offer higher performance but lead to higher cost and latency, and smaller models, which may provide lower performance but run at lower cost.

We run GPT-5, Gemini-2.5-Pro, Claude-4.1-Opus, and DeepSeek-V3.1 using the official APIs from OpenAI, Google, Anthropic, and DeepSeek, respectively. The remaining LLMs are executed locally on a GPU with the following specifications: NVIDIA RTX A6000, 48 GB GDDR6, memory bandwidth of approximately 768 GB/s and peak single-precision (FP32) performance of approximately 38.7 TFLOPS.

\subsection{Evaluation Metrics}

We evaluate LLMs across two dimensions: success rate and latency. 

\textbf{Success Rate}. It measures whether the sequence of actions (tool calls or code) generated by an LLM correctly solves a task according to a given ground truth. To evaluate success rate, we rely on using LLM-as-a-judge, a strategy shown effective in prior work on code evaluation \cite{wang2025can}. 
We used LLM-as-a-judge because multiple valid action sequences may fulfill the same instruction, making direct comparison against a single ground truth misleading. For instance, in a task such as “\textit{Move all objects to the basket}”, different orders of placing the objects are equally correct, even though they differ from the ground truth sequence.

We provide the judge with three inputs: (1) the natural language task, (2) the ground truth action sequence, and (3) the LLM-generated action sequence. Based on these inputs, the judge outputs an integer score of either 0, 1, or 2. Table \ref{tab:scoring_rubric} shows the definition of each score. 

\begin{table}[h]
\centering
\begin{tabular}{cl}
\toprule
\textbf{Score} & \textbf{Description} \\ 
\midrule
0 & The predicted solution does not solve any subtask.. \\
1 & The predicted solution solves at least one subtask. \\
2 & The predicted solution fully solves all subtasks. \\
\bottomrule
\end{tabular}
\caption{Scoring rubric for evaluating predicted solutions.}
\label{tab:scoring_rubric}
\end{table}

To ensure consistency in scoring, we provide explicit evaluation guidelines to the judge. A score of 2 is assigned when all subtasks are fully completed with all required parameter values being respected, ensuring that the correct objects are manipulated in the correct positions. A score of 1 is assigned when at least one subtask is completed correctly, but one or more other subtasks contain errors, are missing, or include irrelevant actions unrelated to the main task. A score of 0 is assigned when no subtask is correctly fulfilled.

We employed three LLMs — GPT-4.1, Claude-Sonnet-4, and Gemini-2.5-Flash — as judges, assigning scores based on the majority vote. If all three LLMs assigned different scores, the final score was computed as the average. These evaluators were chosen to be distinct from the models assessed in the benchmark to minimize potential bias. The temperature was fixed at 0.0 for all models to ensure consistency and reproducibility across runs.



We report the average success rate (AVG) for each LLM, computed across all 56 instructions. In addition, we present a breakdown of results by paraphrase category: Canonical (CAN), Lexical (LEX), Syntactical (SYN), Semantics (SEM), and High-Level Reasoning (HLR).


\textbf{Latency}. It measures the time (in seconds) that an LLM spent to complete a task. We report the average latency across all tasks to provide a single representative value.

\subsection{Evaluation Design}

We design the evaluation process to compare performance across three dimensions: (1) task executor mode (\ac{tap} vs \ac{cap}); (2) LLMs; and (3) paraphrase categories. 

We use a lightweight simulation environment to conduct experiments efficiently. This environment provides APIs for robot movement and perception that return placeholder poses, mimicking the outputs of a real robot or simulator. Robot movements are considered successful as long as the parameters follow the correct format (e.g., a dictionary specifying position and orientation). This setup allows rapid evaluation of generated code and tool calls without compromising the validity of the results, as our primary goal is to assess the quality of the generated actions rather than the detailed robot dynamics. To ensure realism, all ground-truth code and tool calls were validated in the Gazebo simulator using the Interbotix \texttt{wx250s} arm \cite{wx250s_specs}, confirming that actions generated by the framework are executable in a physically realistic environment.



\section{Results}
\label{sec:results}

Table \ref{tab:agent_policy_perf} summarizes the performance of our proposed \ac{alrm} architecture across modes of operation, LLMs, and paraphrasing categories. We refer to this table when describing results in the remainder of this section.
For the purpose of analysis, we categorize LLMs into large-scale models (GPT-5, Gemini-2.5-Pro, Claude-4.1-Opus, DeepSeek-V3.1) and small-scale models (Falcon-H1-7B, Qwen3-8B, Llama-3.1-8B, DeepSeek-R1-7B, Granite-3.3-8B, Mistral-7B), allowing us to compare performance across model scales.

\begin{table*}[h!]
\centering
\caption{Performance of \ac{alrm} framework across modes, LLMs, and paraphrase categories. \textbf{Success Rate} is divided in score counts (0, 1, 2) and per-category percentages (CAN, LEX, SYN, SEM, HLR, AVG). Latency is reported in seconds.}
\label{tab:agent_policy_perf}
\begin{tabular}{llccccccccc c}
\toprule
\multirow{4}{*}{\textbf{Mode}} &
\multirow{4}{*}{\textbf{LLM}} &
\multicolumn{9}{c}{\textbf{Success Rate}} &
\multirow{4}{*}{\textbf{Latency (s)}} \\
\cmidrule(lr){3-11}
 & & \multicolumn{3}{c}{\textbf{Score (counts)}} & \multicolumn{6}{c}{\textbf{Per Category (\%)}} & \\
\cmidrule(lr){3-5} \cmidrule(lr){6-11}
 & & \textbf{0} \ & \textbf{1} & \textbf{2} & \textbf{CAN} & \textbf{LEX} & \textbf{SYN} & \textbf{SEM} & \textbf{HLR} & \textbf{AVG} & \\
\midrule
\multirow{10}{*}{Code-as-Policy}
 & Claude-4.1-Opus & 0 & 8 & 46 & 94.4 & 94.4 & 94.4 & 88.9 & 91.7 & \textbf{92.6} & 33.44 \\
 & GPT-5           & 0 & 10 & 44 & 88.9 & 94.4 & 100.0 & 88.9 & 86.1 & 90.7 & 145.59 \\
 & DeepSeek-V3.1   & 0 & 17 & 37 & 88.9 & 88.9 & 83.3 & 83.3 & 80.6 & 84.3 & 69.83 \\
 & Gemini-2.5-Pro  & 0 & 29 & 25 & 72.2 & 77.8 & 66.7 & 77.8 & 72.2 & 73.1 & 52.62 \\
 \cmidrule(lr){2-12}
 & Falcon-H1-7B    & 0 & 17 & 37 & 83.3 & 88.9 & 88.9 & 88.9 & 77.8 & \textbf{84.3} & 24.89 \\ 
 & Qwen3-8B        & 7 & 24 & 23 & 66.7 & 72.2 & 88.9 & 61.1 & 50.0 & 64.8 & 343.33 \\
 & Llama-3.1-8B    & 2 & 28 & 24 & 72.2 & 77.8 & 77.8 & 77.8 & 58.3 & 68.5 & 20.13 \\
 & Granite-3.3-8B  & 6 & 38 & 10 & 61.1 & 56.6 & 56.6 & 44.4 & 52.8 & 53.7 & 24.48 \\
 & DeepSeek-R1-7B  & 31 & 23 & 0 & 33.3 & 5.6 & 5.6 & 27.8 & 27.8 & 21.3 & 53.55 \\
 & Mistral-7B      & 45 & 9 & 0 & 11.1 & 11.1 & 5.6 & 5.6 & 8.3 & 8.3 & 18.96 \\
\midrule
\multirow{10}{*}{Tool-as-Policy}
 & Claude-4.1-Opus & 0 & 7 & 47 & 94.4 & 94.4 & 94.4 & 94.4 & 91.7 & \textbf{93.5} & 82.60 \\ 
 & GPT-5           & 0 & 16 & 38 & 94.4 & 100.0 & 72.2 & 72.2 & 86.1 & 85.2 & 113.88 \\
 & DeepSeek-V3.1   & 0 & 16 & 38 & 88.9 & 94.4 & 83.3 & 83.3 & 80.6 & 85.2 & 161.73 \\
 & Gemini-2.5-Pro  & 0 & 14 & 40 & 88.9 & 94.4 & 77.8 & 88.9 & 86.1 & 87.0 & 117.43 \\
\cmidrule(lr){2-12}
 & Falcon-H1-7B    & 38 & 16 & 0 & 27.8 & 16.7 & 16.7 & 0.0 & 13.9 & 14.8 & 161.34 \\
 & Qwen3-8B        & 5 & 40 & 9 & 55.6 & 61.1 & 55.6 & 55.6 & 47.2 & \textbf{53.7} & 392.08 \\
 & Llama-3.1-8B    & 50 & 4 & 0 & 5.6 & 0.0 & 0.0 & 5.6 & 5.6 & 3.7 & 36.07 \\
  & Granite-3.3-8B  & 54 & 0 & 0 & 0.0 & 0.0 & 0.0 & 0.0 & 0.0 & 0.0 & 19.07 \\
& DeepSeek-R1-7B\footnotemark & -- & -- & -- & -- & -- & -- & -- & -- & -- & -- \\
 & Mistral-7B      & 54 & 0 & 0 & 0.0 & 0.0 & 0.0 & 0.0 & 0.0 & 0.0 & 10.81 \\ 
\bottomrule
\end{tabular}
\end{table*}


\subsection{Operation Mode Comparison}

We found that \textbf{most large-scale LLMs achieve higher success rates with TaP than CaP, although this leads to increased latency. Small-scale LLMs (up to 8B parameters) struggle with TaP, though some achieve competitive success rates with CaP while maintaining lower latency.}

For large-scale LLMs, Claude-4.1-Opus reached the highest average success rate using TaP (93.5\%), representing a 0.9\% improvement over CaP, while latency increased from 33.44s to 82.60s. Similar trends were observed for DeepSeek-V3.1, and Gemini-2.5-Pro, with improvements of 0.9\% and 13.9\%, respectively, from CaP to TaP. GPT-5 was the exception, with CaP outperforming TaP (90.7\% vs 85.2\%). This demonstrates a trade-off between CaP and TaP: tool-based execution leverages nested reasoning capabilities of large LLMs better to fulfill manipulation tasks but requires more processing time.

Some small-scale LLMs demonstrate impressive performance under CaP. Falcon-H1-7B achieves a 84.3\% success rate and with a latency of 24.89 seconds. Llama-3.1-8B and Qwen3-8B and Llama-3.1-8B also perform reasonably well under CaP, achieving an average success rate of 64.8\% and 68.5\%, respectively. These results highlight that smaller models can  generate executable control code for robotic tasks.

However, small-scale LLMs struggle over TaP, where success rates often drop significantly. Qwen3-8B is an exception, achieving 53.7\% success rate under TaP, although its average task latency increases to 392.08 seconds due to its thinking output. Qualitative inspection indicates that small-scale models often face difficulties handling multiple consecutive nested calls, particularly in pick-and-place operations, leading to partial or failed completions in TaP mode. 

\subsection{LLMs Comparison}

\footnotetext{DeepSeek-R1-7B does not support tool calling.}

We found that \textbf{Claude-4.1-Opus had the best performance in large-scale LLMs with TaP and Falcon-H1-7B had the best performance in small-scale LLMs with CaP.}

Among large-scale LLMs, Claude-4.1-Opus achieved the highest performance in both success rate and latency across TaP and CaP modes. GPT-5 also achieved high success rates, particularly in CaP, but its high latency (145.59s in CaP) limits practical usability. 
Gemini-2.5-Pro and DeepSeek-V3.1 performed similarly, althought Gemini-2.5-Pro showed a slightly better success rate and latency than DeepSeek in TaP. Furthermore, no large scale LLMs received a score of 0, reflecting their robustness across multistep robotic manipulation tasks.

Among small-scale LLMs, Falcon-H1-7B stands out with an impressive 84.3\% success rate in CaP. This is 11.2\% higher than Gemini-2.5-Pro under CaP and equal to DeepSeek-V3.1, despite Falcon-H1-7B having far fewer parameters and one of the lowest processing times (24.89s). This highlights Falcon-H1-7B strong competitiveness relative to large-scale LLMs.


Qwen3-8B shows moderate performance across both modes, but its practicality is limited by high latency (392.08s in TaP and 343.33s in CaP). Llama-3.1-8B also performs relatively well in CaP, achieving a 68.5\% success rate, making it another notable small-scale model. Granite-3.3-8B lags behind, with a 53.7\% success rate and only 10 scores of 2 in CaP.


Other small-scale LLMs, including DeepSeek-R1-7B (21.3\%) and Mistral-7B (8.3\%), perform poorly under CaP. Errors mostly occurred because the models struggled to generate meaningful actions during planning. For example, Mistral-7B often failed to generate a single action per turn, generating instead the entire plan with the final answer in a single shot. This behavior prevented both CaP and TaP from generating actions and, consequently, resulted in low performance.

\subsection{Task Categories}

We found that \textbf{large-scale LLMs perform consistently well across all task categories, particularly in the lexical one, while small-scale LLMs in CaP mode excel in the syntactical category but struggle with high-level reasoning.}


For large-scale LLMs across both CaP and TaP, the canonical category had higher success rate  (average of 88.8\% among models), showcasing their ability to fulfill direct requests, with GPT-5 and Claude-Opus-4.1 achieving both 94.4\% under TaP. Lexical tasks had the highest success rate (average of 92.3\%), showing strong performance of LLMs in understanding diverse wordings and synonyms for object names; GPT-5 scored perfectly under TaP, while Gemini achieved only 77.8\% under CaP. High level reasoning (average of 84.3\%), semantics (average of 84.7\%), and syntactic tasks (average of 84.0\%) performed similarly, reflecting standard performance in large-scale LLMs to (1) reason over the task and manipulate objects without their explicit mention; (2) understand informal phrasing and noisy input; and (3) understand different syntax patterns. This suggests that large-scale LLMs struggle more with task difficulty than with natural language variation.



For small-scale LLMs, we analyze the results over CaP. Canonical tasks achieved the highest success rate (average of 54.6\%), showing the ability of small-scale LLMs to fulfill standard imperative requests. Syntactical tasks followed closely (average of 53.7\%), with both Falcon-H1-7B and Qwen3-8B reaching 88.9\% success rate. Lexical and semantics tasks performed similarly (average of 51\%), with Falcon-H1-7B, Llama-3.1-8B, and Qwen3-8B as the top three models. Finally, high-level reasoning tasks had the weakest performance (average of 45.8\%), being the lowest success rates for Falcon-H1-7B, Llama-3.1-8B, and Qwen3-8B models, and highlighting the challenges small-scale models face when reasoning over input to define actions and generate corresponding code.

\section{Conclusion}
\label{sec:conclusion}

In this paper, we proposed \ac{alrm}, a framework for robotic manipulation where LLM agents generate actions via either executable code (\ac{cap}) or tool calls (\ac{tap}). We also proposed a benchmark of linguistic diverse and high-level multistep tasks to evaluate the performance of \ac{alrm} using multiple open-source and closed-source LLMs. Our findings show that closed-source LLMs achieve higher performance and increased latency following \ac{tap} operation mode, with Claude-4.1-Opus achieving the best results. Small-scale open-source LLMs perform well in generating executable robot code, with Falcon-H1-7B surpassing Gemini-2.5-Pro and matching DeepSeek-V3.1 while maintaining much lower latency. However, these open-source models struggle with tool-based execution for high-level reasoning tasks.
Based on these results, we recommend Claude-4.1-Opus for API use, Falcon-H1-7B for local execution, and the development of small-scale LLMs with enhanced nested tool-calling abilities.




Finally, this study has limitations that also suggest opportunities for future research. First, the performance of the \ac{alrm} framework depends on the prompts used for both task planning and execution. The results presented may change across LLMs depending on the prompt, motivating future work to evaluate our framework with different prompts. Second, the benchmark is limited to three environments and mainly pick-and-place tasks; expanding to more environments and task types, including navigation, would provide broader evaluation. Third, while our experiments validate the framework in simulation, integrating real robots and perception components would offer a clearer understanding of its practical performance.




\bibliographystyle{IEEEtran}
\bibliography{biblio}

\end{document}